\documentclass{amia}
\usepackage{lipsum} %Remove if not needed
\graphicspath{{./figures}}
\usepackage{hyperref}
\usepackage{algorithm}
\usepackage{algpseudocode}

\usepackage{amsmath}
\usepackage{algpseudocode}
\usepackage{booktabs}

\begin{document}

\title{\textit{Active Learning for Forecasting Severity among Patients with Post Acute Sequelae of SARS-CoV-2}}

\author{Jing Wang, PhD$^1$, Amar Sra, MD, MS$^2$, Jeremy C. Weiss, MD, PhD$^1$ }

\institutes{
    $^1$ National Library of Medicine, Bethesda, MD \\$^2$The George Washington University, Washington, DC
}

\maketitle

\section*{Abstract}

The long-term effects of Postacute Sequelae of SARS-CoV-2, known as PASC, pose a significant challenge to healthcare systems worldwide. Accurate identification of progression events—such as hospitalization and reinfection—is essential for effective patient management and resource allocation. However, traditional models trained on structured data struggle to capture the nuanced progression of PASC. In this study, we introduce the first publicly available cohort of 18 PASC patients, with text time series features based on Large Language Model Llama-3.1-70B-Instruct and clinical risk annotated by clinical expert \cite{touvron2023llama}. We propose an Active Attention Network to predict the clinical risk and identify progression events related to the risk. By integrating human expertise with active learning, we aim to enhance clinical risk prediction accuracy and enable progression events identification with fewer number of annotation.  The ultimate goal is to improves patient care and decision-making for  SARS-CoV-2 patient.

\section*{Introduction}

Long COVID is an often debilitating illness that occurs in at least 10\% of severe acute respiratory syndrome coronavirus 2 (SARS-CoV-2, known as PASC) infections. There are 65 million individuals worldwide are estimated to have long COVID until 2023 \cite{davis2023long}. Long COVID symptoms can be severe, such as more than 200 symptoms have been identified with impacts on multiple organ systems. Long Covid remains an unsolved, complex, and urgent healthcare crisis which has affected one in nine adults in United States who have ever had COVID-19, according to CDC. Hence PASC poses a significant challenge to healthcare systems worldwide. The research of PASC to enhance recovery initiative requires intensive attention.

In this work, we study the clinical risk of patients with PASC infections from case reports. We release a dataset with 18 PASC infected patients with clinical expert annotated clinical risk level and Large Language Model Llama-3.1-70B-Instruct generated structured clinical event-time series. The clinical risk level is defined as Table \ref{tab:risk_level}.
\vspace{0.1in}
\begin{table}[h]
	\centering
		\caption{Risk Level Definitions}
	\begin{tabular}{|l|l|}
		\hline
		\textbf{Risk level} & \textbf{Definition}                                             \\ \hline
		Low                & Symptoms with low or some burden on quality of life            \\ \hline
		High               & Requiring hospitalization, ICU stay or death                   \\ \hline
	\end{tabular}
	\vspace{0.2in}
	\label{tab:risk_level}
\end{table}\vspace{-0.2in}
\begin{table}[h!]
	\centering
		\caption{Example of generated clinical events and corresponding timestamp.}
	\begin{tabular}{|l|l|}
		\hline
		\textbf{Event}                          & \textbf{Timestamp} \\ \hline
		%		anxiety                                 & -672               \\ \hline
		depression                              & -672               \\ \hline
		%		insomnia                                & -672               \\ \hline
		mild cognitive impairment               & -672               \\ \hline
		%		fever                                   & -672               \\ \hline
		%		cough                                   & -672               \\ \hline
		%		off-white sputum                        & -672               \\ \hline
		mild depression                         & -240               \\ \hline
		%		mild insomnia                           & -240               \\ \hline
		mild brain fog                          & -240               \\ \hline
		%		30 years old                            & 0                  \\ \hline
		female                                  & 0                  \\ \hline
		TBS sessions                            & 24                 \\ \hline
		%		improved symptoms                       & 120                \\ \hline
		%		BAI score improved                      & 120                \\ \hline
		BDI score improved                      & 120                \\ \hline
		%		HAMD score improved                     & 120                \\ \hline
		3-month follow-up                       & 744                \\ \hline
		%		clinical improvement continued          & 744                \\ \hline
	\end{tabular}
	\vspace{0.1in}
	\label{tab:events}
\end{table}
The clinical timestamp is based on admission event set as time 0, then the events that happened before admission have negative timestamp, while the events after admission is with positive timestamp. We convert all the time annotation in hours. Table \ref{tab:events} shows an example of annotated clinical event-time series for case report PMC10077184. 

Given the clinical event-time series, we developed an Active Attention model for risk prediction. We include active learning in the risk prediction model training, with the hope is that a good risk prediction model can be learned with significantly fewer labels by actively directing the queries to informative samples. It is validated by our experimental results as shown in Figure \ref{fig:test4} and Figure \ref{fig:test7}. The framework is composed of an Attention layer for embedding of events, an uncertainty-based active learning to select most informative samples, a Multi-Layer Perceptron layer to produce the probability distribution over the classes. The framework simulates the human-in-the-loop annotation process, that is in each iteration, we apply uncertainty based active learning to select the most informative sample and query its label, then the labeled sample is combined with existing training data to retrain the risk prediction model. The uncertainty of unlabeled sample is based on margin of two class probabilities. The samples with smallest margin is considered with most informative for the model. The intution is that the samples with smaller margin is closer to the decision boundary. Hence training the moddl on such data  helps improves the performance of model.

The simulation performance on the real-world annotated data shows that active learning consistently outperforms random selection given the same number of labeled samples. The model equipped with active learning also reaches the desired accuracy with few number of labeled samples. To identify clinical events more related to the risk level, we use Attention layer to compute the important score for clinical events and related timestamps. The result shows that the model with active learning is better at identifying treatment plans, medications, and demographic information that can influence the immediate risk of patient. 

\textbf{Our Contributions:} We present and analyze a comprehensive framework for Active Attention Network-based learning specifically designed to address the challenges of risk prediction and progression event detection in a novel Long COVID dataset. Our contributions include:
\begin{itemize}
	\item Long COVID dataset collection and annotation:
	We curate and annotate a structured dataset containing detailed risk levels and event-time series, offering a valuable resource for analyzing disease progression and risk assessment.
	\item Active Attention Network development:
	We propose an Active Attention Network (AAN) that effectively learns from the annotated dataset to perform both risk prediction and important progression event detection. The model identifies key features contributing to risk assessment, enhancing interpretability and robustness.
	\item Active learning for improved efficiency:
	By incorporating active learning techniques, our model achieves high accuracy with significantly fewer labeled samples. Unlike standard approach such as random selection, our method selects the most informative samples, improving both sample efficiency and prediction quality. The results are shown in Figure \ref{fig:test4} and Figure \ref{fig:test7}. Active Attention Network demonstrates its ability to select features most closely related to risk. Some important features are shown in Table \ref{tab:events}. It helps understand the risk factors of long COVID.
	\item Long COVID and the Recover Initiative: The released dataset and accompanying analysis provide critical insights into the progression of Long COVID, elucidating its long-term effects on various organ systems. The clinical event-time dataset contribute to a more comprehensive understanding of disease mechanisms. The identified important features by AAN supports the development of evidence-based treatment recommendations aimed at enhancing recovery efforts under the RECOVER Initiative, thereby improving patient outcomes and guiding future clinical interventions.
	 
\end{itemize}

\section*{Related works}

\noindent{\bf Long COVID} also referred to as ``post-acute sequelae of COVID-19'', is a multisystem inflammatory syndrome consisted of severe symptoms that 3 months follow a severe acute respiratory syndrome coronavirus 2 infection. There are at least 65 million individuals around the world have long COVID based on more than 651 million documented COVID-19 cases worldwide \cite{ballering2022persistence}. There are many challenges relating to pathophysiology, effective treatments and risk factors \cite{fair2022patients,zang2023data}. Risk factors of PASC include female sex, type 2 diabetes, the presence of specific autoantibodies \cite{su2022multiple}, connective tissue disorders \cite{renz2022pathobiology}, and a third of people with long COVID have no identified pre-existing conditions. We follow this line and develop a risk prediction model based on an Active Attention Network that considers the temporal information of the disease propagation. The conditions of long COVID patients are not well documented in electronic healthcare records because of lacking post-viral knowledge and misinformation and imperfect code. We release a dataset of long COVID patients with clinical event-time annotated from case reports. As shown in Table \ref{tab:comparison}, Our model effectively identifies comprehensive and detailed features associated with clinical risk, such as age and symptoms like``extreme shortness of breath''.

\noindent{\bf Acitve Learning} has been studied for decases to improve the model performance using small sets of labeled data, significantly reducing the annotation cost \cite{cohn1996active,balcan2007margin}. There are two primary categories of AL sampling strategies: diversity-based \cite{sener2018active} and uncertainty-based \cite{liang2020alice,shen2022metric,tmlr2022}. The diversity-based methods select the samples that can best represent the dataset distribution. Uncertainty based methods focus on selecting the samples with high uncertainty, which is due to the data generation or the model process, such as Margin-based active learning \cite{balcan2007margin}, Bayesian Active Learning by Disagreements \cite{gal2017deep}. This work use the uncertainty-based method based on margin of class distribution. It is effective for binary classification as our case. In each iteration, we identify and select the most informative samples based on the margin from the unlabeled pool iteratively. 

\section*{Clinial Event-Time Sequences Annotated by Large Language Model}

 Our dataset came from the PubMed Open Access subset (PMOA) repository as of 2024/04/18 comprising 1,499,346 full-text articles. We use the following regular expressions to identify clinical reports with PASC infection: (1) presence of ``case report'' or ``case presentation'', (2) presence of the string ``year-old'' or ``year old'', and (3) presence of the string ``long covid'' or ``long-covid''. It leads to a subset with 93 case reports. Then we use Large Language Model Llama-3.1-70B-Instruct to identify the number of cases in each report and identify case reports with long COVID patients based on multiple terminologies associated with the condition, including chronic fatigue, post-traumatic stress disorder, brain fog, shortness of breath, chest pain, fatigue, post-exertional malaise, dyspnea, shortness of breath,
 chest discomfort, Cough, PTSD, anxiety, depression, impaired memory, muscle pain, myalgias, poor concentration, insomnia, anosmia, loss of smell, persistent symptoms, post-acute sequelae. Based on the revised prompt in \cite{summit25}, LLM selects 43 reports containing one case and generates their clincal event-time sequences. The clinically trained expert manually reviews the 43 reports and identifiest that there are 18 reports about patients with PASC infection.

\section*{Active Attention Network for Risk Prediction}

Given the clinical event-time sequences
 \[
\left\{ \langle e_i, t_i \rangle _{i=1}^{n_j}, y_j\right\}, \quad \text{where} \quad j \in [0, N],
\]
where $e_i$ is the text clinical event, $t_i$ is the timestamp of the event $e_i$ in hours, $y_i\in [0,1]$ is the risk level, $n_j$ is number of event-time for $j$th report, there are $N$ reports in total. We develop an Active Attention Network for the clinical risk prediction. Our model is a neural network architecture designed to incorporate per-feature attention mechanisms to enhance the learning process. The architecture is structured as in Table \ref{tab:attentionmlp}.

\begin{table}[h]
	\caption{Architecture of Attention Attention Network}
%	\vspace{0.1in}
	\centering
	\begin{tabular}{|c|c|}
		\hline
		\textbf{Network Component} & \textbf{Equation} \\
		\hline
		Attention Mechanism & 
		$\begin{array}{c}
		\boldsymbol{a} = \sigma(\mathbf{W}_{\text{attn}} \boldsymbol{x}) \\
		\boldsymbol{z} = \boldsymbol{a} \odot \boldsymbol{x}
	\end{array}$ \\	 
		\hline
		Feedforward Layer 1 & 
		$\begin{array}{c}
			\boldsymbol{h}_1 =  \text{ReLU} ( \text{BatchNorm} (\mathbf{W}_2 \boldsymbol{z} + \boldsymbol{b}_1)) \\
			\boldsymbol{h}_1 = \text{Dropout}(\boldsymbol{h}_{1, \text{temp}})
		\end{array}$ \\
		\hline
		Feedforward Layer 2 & 
		$\begin{array}{c}
			\boldsymbol{h}_{2, \text{temp}} = \text{ReLU}(\text{BatchNorm}(\mathbf{W}_2 \boldsymbol{h}_1 + \boldsymbol{b}_2)) \\
			\boldsymbol{h}_2 = \text{Dropout}(\boldsymbol{h}_{2, \text{temp}})
		\end{array}$ \\
		\hline
		Output Layer & $\boldsymbol{y} = \text{Softmax}(\mathbf{W}_{\text{out}} \boldsymbol{h}_2 + \boldsymbol{b}_{\text{out}})$ \\
		\hline
%		Loss Function & $\mathcal{L}_{\text{CE}} = - \frac{1}{N} \sum_{i=1}^{N} \sum_{j=1}^{C} y_{ij} \log(\hat{y}_{ij})$ \\
%		\hline
	\end{tabular}
	\label{tab:attentionmlp}
\end{table}

The model employs a linear layer to learn attention weights for each input feature. Specifically, the attention mechanism applies a sigmoid activation function to produce weights between 0 and 1, indicating the importance of each feature, where $x$  is the input vector, $\mathbf{W}_{attn}$ is the attention weight matrix, and $\sigma$  denotes the sigmoid activation function.  The resulting attention weights  are applied element-wise to the input features.

There are two feedforward layers. The attention-adjusted features  are passed through two fully connected layers with ReLU activations, batch normalization, and dropout regularization, where $W_1$  and $W_2$ are the weight matrices, and $b_1$ and  $b_2$ are the biases.

The output layer applies a linear transformation followed by a softmax function to produce class probabilities. The model is trained using the cross-entropy loss function, which is defined as:
\begin{equation}
	\text{Loss} = - \frac{1}{N} \sum_{j=1}^{N} \sum_{c=1}^{C} y_{cj} \log(\hat{y}_{cj})
\end{equation}
where  $N$ is the number of training samples, $C=2$  is the number of classes, $y_i$  is the true label (1 if the sample belongs to class 1, 0 otherwise), and $\hat{y}$ is the predicted probability for class .

\noindent{\bf Active Learning}. In the proposed framework, we employ an uncertainty-based active learning strategy to selectively query the most informative samples from the unlabeled pool. The underlying intuition is that the model is most uncertain about samples that are closest to the decision boundary. Therefore, querying such samples can significantly improve the model’s performance. 

For each unlabeled sample $\langle e_j, t_j \rangle$ in the unlabled pool, the current Active Attention model produces a probability distribution over the classes:
\[
\boldsymbol{p} = f(w, \langle e_j, t_j \rangle)
\]
where \( \boldsymbol{p} = [p_0, p_1] \) represents the probabilities assigned to the two classes. The \textbf{uncertainty score} for each sample is computed as the absolute difference between the two probabilities:
\[
s(\langle e_j, t_j \rangle) = |p_0 - p_1|
\]

The sample with the \textbf{lowest uncertainty score} is selected for annotation. This corresponds to:
\[
\text{Selected Index} = \arg \min_j \left( |p_{0j} - p_{1j}| \right)
\]

\begin{algorithm}
	\caption{Uncertainty-Based Active Attention Network Learning}
	\begin{algorithmic}[1]
		\State \textbf{Input:} a set of unlabeled instances  $ U_0 = \{\boldsymbol{\langle e_i, t_i\rangle}_{i=1}^{j} \}^p_{j=1}$, a set of labeled instances $L_0 = \{\boldsymbol{\langle e_i, t_i\rangle}_{i=1}^{j},y_{j} \}^l_{j=1}$, initial Active Attention Network $f(w_0)$, number of iterations $T=p$.
		\For{$t = 1, \cdots, T$} 
		\State $U_t \longleftarrow U_{t-1}$
		\For{$\{\boldsymbol{\langle e_i, t_i\rangle}_{i=1}^j \in  U_t$}
		\State Compute probability distribution \( \boldsymbol{p}_j = f(w_{t-1}, \boldsymbol{\langle e_i, t_i\rangle}_{i=1}^j) \)
		\State Calculate uncertainty score \( s(\boldsymbol{\langle e_i, t_i\rangle}_{i=1}^j) = |p_{0j} - p_{1j}| \)
		\EndFor
		\State Select most informative example $\boldsymbol{\langle e_i, t_i\rangle }_{i=1}^* \longleftarrow \arg \min_j  s(\boldsymbol{\langle e_i, t_i\rangle}_{i=1}^j) $, $\forall \boldsymbol{\langle e_i, t_i\rangle}_{i=1}^j \in U_t$
		\State Query the label of $y_*\longleftarrow \boldsymbol{\langle e_i, t_i\rangle}_{i=1}^*$
		\State Update training dataset $L_t\longleftarrow L_{t-1} \cup \{\boldsymbol{\langle e_i, t_i\rangle}_{i=1}^*\}$
		\State Update Attention Attention Network $w_t\longleftarrow f(w_{t-1}|L_t)$
		\State Update unlabeled dataset $U_t\longleftarrow U_t/ \{\boldsymbol{\langle e_i, t_i\rangle}_{i=1}^*\}$
		\EndFor
	\end{algorithmic}
	\label{alg}
\end{algorithm}

We query the label of selected sample and use it to update the Active Attention Network. The unlabeled pool and labeled dataset are updated accordingly. The main algorithm with all iterations is introduced in Algorithm \ref{alg}. %The random selection picks a random unlabeled sample from $U_i$ and queries its label, elimiating it from unlabeled set, and udpates the model at each Round $t$.

\section*{Experiments}

\noindent{\bf Feature Embedding}
We learn the embedding of clinical events by ``NeuML/pubmedbert-base-embeddings'' \cite{gu2021domain}. Each clinical event is converted to a 768 dimension vector, then we reduce the feature space to 32 with linear projection. The maximal number of event-time pairs for one case is set as 150. The timestamp is normalized and combined with the 32 event feature space. The total dimension of the feature space for one case report is 4,950. The features are fed to the Active Attention Network for training.

We select $n_\text{test}$ samples for testing, and $n_\textit{train}$ samples without overlap. The remaining samples are allocated to an unlabeled pool for subsequent selection. Our experimental setup includes testing with $n_\text{test}=[5,7], n_\text{train}=4$. For each data split, we repeat the process five times to obtain robust accuracy measurements. 
\begin{figure}[H]
	\centering
	\includegraphics[width=1.0\textwidth]{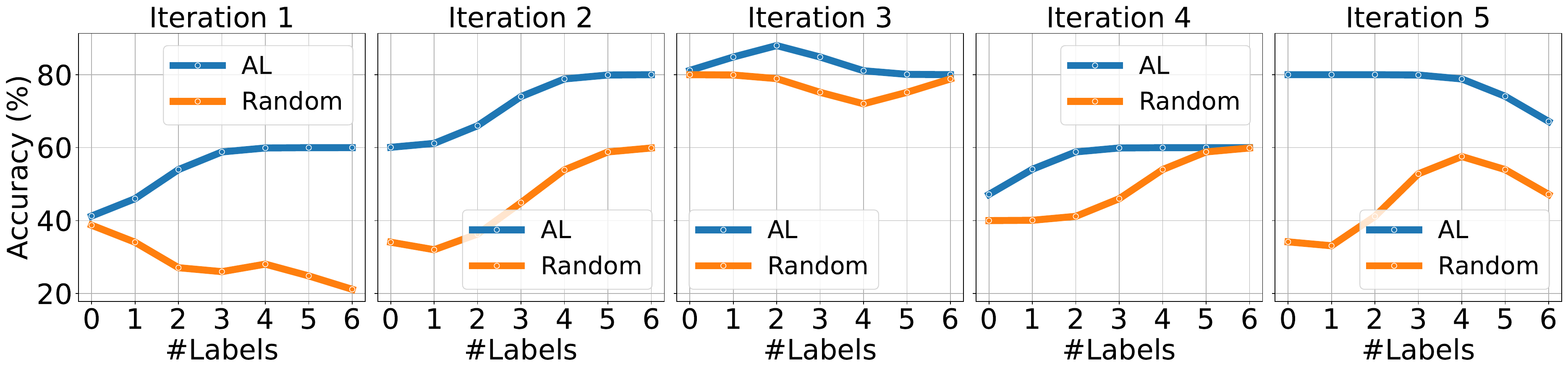}
	\caption{Comparison of AL and Random Strategies over 5 Random Data Splits on Testing Dataset with size 5. }
	\label{fig:test4}
\end{figure}
\vspace{0.3in}
\begin{figure}[H]
	\centering
	\includegraphics[width=1.0\textwidth]{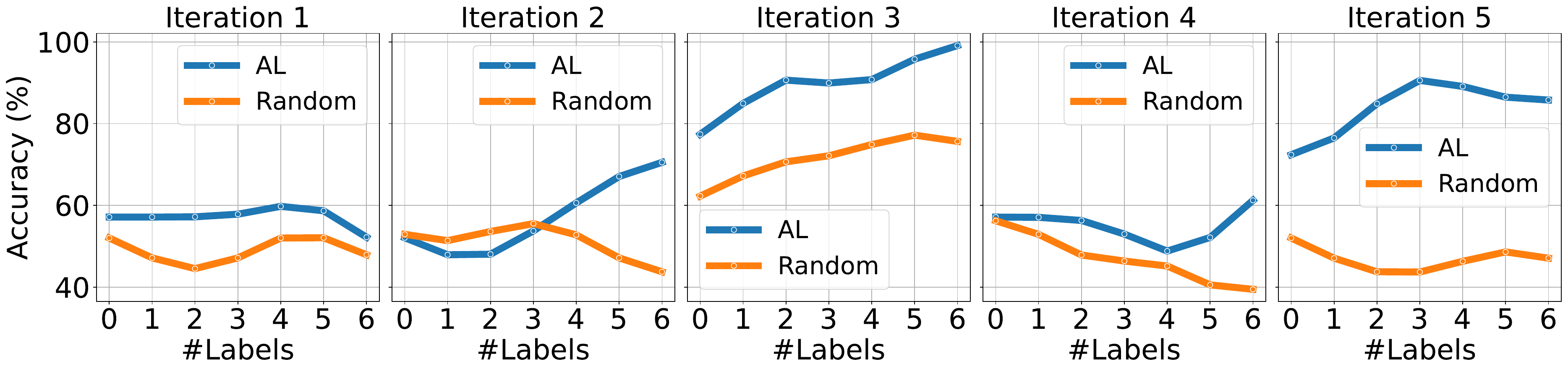}
	\caption{Comparison of AL and Random Strategies over 5 Random Data Splits on Testing Dataset with size 7. }
	\label{fig:test7}
\end{figure}
\noindent{\bf{Comparison of Model Performance with AL and Random Sampling}}. We compare the performance of Active Attention Network with AL and Random Sampling. The results are shown in Figure \ref{fig:test4} with testing data size 4 and Figure \ref{fig:test7} with testing data size 7.
For each subplot (Iteration 1 to Iteration 5), we randomly split the data into train and test five times and evaluate the performance of AL and Random selection strategies.
The curvy lines show the smoothed accuracy progression over iterations, where each subplot indicates the results of one random split.  The X-axis represents the increase in training samples during iterative training. The model with active learning consistently outperforms randsom selection most of times.

\noindent{\bf Important Features For Risk Prediction}

\begin{table}[H]
	\centering
		\caption{Comparison between Active Learning and Random Selection in Identifying Relevant Features for Risk Assessment}
	\begin{tabular}{|l|l|p{4.0cm}|p{3.5cm}|p{3.5cm}|}
		\hline
		\textbf{Case Report} & \textbf{Risk} & \textbf{Common Features} & \textbf{Active Learning (AL) Unique Features} & \textbf{Random Selection (RS) Unique Features} \\ \hline
		%		PMC10008181 & 1 (High) & atorvastatin 20 mg, involving the proximal left middle cerebral artery territory, midline shift of 0.5 cm, normal platelet count & normal leukocyte count & significant edema \\ \hline
		PMC9451509 & 0 (Low) & Caucasian, discharged on  prednisolone taper, no travel outside of portugal& elevated erythrocyte, sedimentation rate, female & normocytic anaemia, proinflammatory state \\ \hline
		PMC10129030 & 1 (High) & N1 gene 0 copies/$\mu$L, Unknown Event 147, bronchiolar metaplasia & None (Identical to RS) & None (Identical to AL) \\ \hline
		PMC10173208 & 1 (High) & body mass index 23.9 kg/m2, no improvement & bronchoscopy, leukocytosis & SpO2 at 93\% on room air, referred to our hospital \\ \hline
		PMC10469423 & 0 (Low) & SEM, Time, aspirin & genotype VI.2.1.1.2.2, long COVID & antipyretic agents \\ \hline
		%		PMC8405236 & 1 (High) & extreme shortness of breath, laboratory and biomarker testing, moved into the ICU & Dexamethasone, at work & bilateral pneumonia, prescribed daily evening nebulizer treatments \\ \hline
		PMC8958594 & 0 (Low) & admitted to hospital, diagnosed with COVID-19-related encephalopathy & 62 years old, experienced cataracts, started taking Neurontin/gabapentin & brain CT/MRI, experienced vague visual hallucinations, performed poorly on SHAPS \\ \hline
		PMC9066079 & 0 (Low) & Time, admitted to the hospital, chest computed tomography, physical examination & treatment with codeine & gabapentin increased to 300 mg bid \\ \hline
		PMC9514285 & 1 (High) & Unknown, acetazolamide discontinued, grade III edema & IV thiamine 200 mg every 12 hours & grade II edema \\ \hline
		PMC10077184 & 0 (Low) &Pittsburgh Sleep Quality Index (PSQI), WMS score improved, chest X-ray, fenoterol 2.5 mg/d & female & mild anxiety \\ \hline
		PMC8405236 & 1 (High) & extreme shortness of breath, laboratory and biomarker testin, moved into the ICU, laboratory and biomarker testing & Dexamethasone at work & bilateral pneumonia, prescribed daily evening nebulizer treatments \\ \hline 
		
	\end{tabular}
	\vspace{0.3in}
	\label{tab:comparison}
\end{table}

We show the top 5 features selected by the model with Active Learning (AL) and model with Random Selection (RS) when training with a dataset with 11 samples in Table \ref{tab:comparison}. We selecte case reports belonging to both high risk and low risk category. There are 4 case reports about patient with low risk, and 3 case reports about patients with high risk. The selected features are either clinical event, ``Time'' (timestamp) or ``Unknown''.  When the time annotation is selected as a important feature, we use ``Time'' in Table \ref{tab:comparison}. We set maximal number of clinical event, timestamp pair with 150 for one case report, if there are no more than 150 features, we use 0 filling. If the selected feature fall to the 0 filling, the important feature is ``Unknown''. Both models select important features related to risk level, such as ``diagnosed with COVID-19-related encephalopathy'', ``chest computed tomography'',  as shown in ``Common feature'' column.

To compute the feature importance score, we select the attention weight layer of the Active Attention network as shown in Table \ref{tab:attentionmlp}, and multiply with the feature embedding of the training data as feature importance score, we rank the scores and map the embedding to original features, that is event, timestamp or ``Unknown''. There are 5 columns in Tabl \ref{tab:attentionmlp}. The first column ``case report'' is the report id, ``Risk'' shows the clinical expert annotated risk level of the patient in the report, ``Common Features'' is about important features selected by the model with active learning and random selection, ``Active Learning (AL) Unique Features'' is about important features selected by Active Attention Network with active learning, ``Random Selection (RS) Unique Features'' shows important features selected by Active Attention model with random selection.

As shown in Table  \ref{tab:comparison}, AL is more accurate at predicting risk factors for patients with post acute sequelae of COVID-19 (PASC). 
\begin{itemize}
	\item AL selects more discriminative features on PMC10173208, PMC10469423, PMC9451509, AND PMC8958594. For example, PMC10173208, AL focuses on infection indicators (``leukocytosis''), while RS highlights respiratory issues (``SpO2 level''). Hence, ``Leukocytosis'' is more related with the risk level. For PMC10469423, AL captures broader chronic conditions (``long COVID''), while RS focuses on acute issues (``antipyretic agents''). ``long COVID'' is a better predictive feature for risk level.  For PMC9451509, AL identifies ``elevated erythrocyte'' which is more related to risk level. For PMC8958594, AL provides better context by including age (``62''), which is a critical factor in assessing risk. 
	\item RS selectes more discriminative features on PMC10077184. Specifically, ``mild anxiety'' is more important feature selected by RS than sex ``female'' selected by AL. 
	\item AL and RS select equally important features on PMC8405236, PMC9066079, and PMC9514285. That is, for PMC8405236, though AL and RS select different events, ``pneumonia''  and ``dexamethasone'', they are equally discriminative since ``pneumonia'' is typically treated in an inpatient setting, and ``dexamethasone'' is commonly administered as part of inpatient care. 
\end{itemize}

In summary, the results show that AL and RS capture different types of features for identifying patient risk. AL is generally better at identifying treatment plans, medications, and demographic information that can influence treatment strategies which is related to critical conditions that indicate immediate risk.

\subsection*{Conclusion}

Long covid symptoms can be surprisingly severe and often require hospital visits and occasionally ICU admissions and death. For those patients not meeting severity for hospital admission, symptoms are often persistent and significantly affect quality of life. Patients meet the definition of long COVID when they have a separate identifiable disease that was likely triggered by an original COVID-19 infection. In this work, we understand the progration of the long COVID patients by first releasing a dataset with long COVID. Our Active Attention Network with the human-in loop annotation results show that the model with active learning significantly outperforms random selection. The selected features by the model are more predictive for risk level evaluation. The findings help enhance Recovery (RECOVER) initiative and answer critical questions about long COVID. 

Our system demonstrates the construction of better models, more label-efficient by 50\%, than models based on random acquisition methods that may be used as more PASC case reports become available.  Using our approach will enable more efficient modeling and understanding as the research community continues to further our understanding the risks of PASC.

This resea­rch was suppo­rted by the Divis­ion of Intra­mural Resea­rch of the Natio­nal Libra­ry of Medic­ine (NLM)­, Natio­nal Insti­tutes of Healt­h. This work utilized the computational resources of the NIH HPC Biowulf cluster \footnote{\href{https://hpc.nih.gov}{Biowulf cluster}}.

%\subparagraph{Acknowledgments}\lipsum[1]

% References as numbers
%\makeatletter
%\renewcommand{\@biblabel}[1]{\hfill #1.}
%\makeatother

% unstr is used to keep citation order
\bibliographystyle{vancouver}
\bibliography{amia}  

\end{document}